# Representation Learning for Medical Data

Karol Antczak
Military University of Technology in Warsaw
Institute of Computer and Information Systems

**ABSTRACT**

We propose a representation learning framework for medical diagnosis domain. It is based on heterogeneous network-based model of diagnostic data as well as modified metapath2vec algorithm for learning latent node representation. We compare the proposed algorithm with other representation learning methods in two practical case studies: symptom/disease classification and disease prediction. We observe a significant performance boost in these task resulting from learning representations of domain data in a form of heterogeneous network.

**1. INTRODUCTION**

Representation learning is a group of machine learning methods that aims to find useful representations of the data. The "usefulness" is typically understood in terms of extraction of features that are meaningful from the point of view of target objective. For neural networks, such representation is defined as a mapping $f$ of input representations to $d$ – dimensional vector space: $f : V \rightarrow R^d$. The development of representation learning is motivated by numerous experimental results showing that extracting the features of the data improves the performance of the network compared to the "naive" data encoding schemes such as binary or one-hot encoding. This is further encouraged by the observations that many deep learning architectures seem to naturally learn the layer-wise representation of the features during the training – a phenomenon which some researchers point out as an important factor contributing to great performance of DL methods. Not without the significance is also the fact that the such internal representations can be, at least in some cases, interpreted by humans, which is a step toward improving the explainability of deep neural models.

To the date, machine learning applications for medical diagnosis did not utilized representation learning, relying on either non-neural feature extraction methods or naive encoding schemes. On the other hand, other deep learning models are used extensively. The primary motivation of our research was to introduce RL into to diagnosis area, encouraged by performance boost observed in other fields. However, this requires to define a data model that can be utilized by representation learning framework. Thus, we propose a formal model of diagnostic data, by means of a heterogeneous network, allowing us to use modern representation learning algorithms for graph-like structures, such as node2vec and metapath2vec. We also develop an extension of metapath2vec that further improves quality of learned representations.

The structure of this paper is as follows. In chapter 2 we briefly present current research status regarding representation learning for networks. Chapter 3 contains proposed model of diagnostic data and representation learning framework for such data. In chapter 4 we present experimental results. Obtained results are discussed in chapter 5.

Implementation of all algorithms used for this paper is available on GitHub: https://github.com/KarolAntczak/multimetapath2vec.

**2. RELATED RESEARCH**

A major milestone in the representation learning field was development of word2vec algorithm which finds efficient representations of words in text corpuses [1]. The basic idea of word2vec was soon applied to other types of data, including networks. This resulted in development of algorithms such as DeepWalk [2] and its generalization, node2vec [3]. Both of these algorithms are designed to learn representations of nodes $V$ given a graph $G=(V,E)$. If input nodes are encoded by one-hot vectors, then the target mapping is a linear transformation and can be represented by the transformation matrix $A$. The learning process is then an optimization task that finds the matrix which, given feature representation of another node, maximizes log-probability of observing a certain "neighbor" node. In other words, we aim to nodes with similar neighbors to have similar feature representation. Since it is ineffective to compute target function for each possible pair of nodes, they are sampled from the network using random walks instead. The general algorithm of node representation learning is given below.

*Algorithm 1: General node representation learning algorithm.*

**Input**: Network $G=(V,E)$, Dimensions $d$, Walks per node $r$, Walk length $l$, Context size $k$
**Output**: node embedding matrix $A \in \mathbb{R}^{|V| \times d}$

Initialize *walks* to empty
**for** $i=1 \to r$ **do**
  **for** $v \in V$ **do**
    *walk* = RandomWalk($G, v, l$)
    Append *walk* to *walks*
  **end**
**end**
$A$ = StochasticGradientDescent($k, d, walks$)
**return** $A$

The exact algorithm of random walks differs between the algorithms. Authors of node2vec note that there are two distinct kinds of node similarities: homophily (occurring in nodes that are close to each other) and structural equivalence (occurring in nodes that have similar structural roles in the network but are not necessarily closely interconnected). The random walk procedure used in word2vec is characterized by two hyperparameters that incorporate both notions of similarity. The unnormalized transition probability from node $v^i$ to node $v^{i+1}$ given previous node $v^{i-1}$ is:

$$\alpha(v^{i+1}|v^i) = \begin{cases} \frac{1}{p} & if\ d(v^{i-1},v^{i+1})=0 \\ 1 & if\ d(v^{i-1},v^{i+1})=1 \\ \frac{1}{q} & if\ d(v^{i-1},v^{i+1})=2 \end{cases} \quad (1)$$

The $d(v^{i-1},v^{i+1})$ denotes the shortest path between previous and next node. The return parameter $p$ controls the likelihood of returning to already visited node, while the in-out parameter $q$ controls the tendency to explore outward nodes. Thanks to this, the random walk can result in different pairs of neighbors, depending on which similarity seems more suitable to the target task. Specifically, sampling strategy used in DeepWalk is the one where $p=1$ and $q=1$, meaning that each node has the same probability of being visited.

Metapath2vec is a modification of node2vec for heterogeneous networks. Heterogeneous network is defined as a graph $G=(V,E,T)$ in which each node and each edge is associated with mapping functions $\phi(v):V \to T_v$, $\phi(e):V \to T_e$, respectively. Instead of using random walks scheme with explicit $p$ and $q$ parameters, metapath2vec utilizes additional information about node types to provide an alternative method, called meta-path-based random walks. The flow of the walk is determined by the so-called meta-path defined as $P:V_1 \xrightarrow{R_1} V_2 \xrightarrow{R_2} ... V_t \xrightarrow{R_t} V_{t+1} ... V_l$ wherein $V_1 ... V_l$ are respective node types and $R_1 \ldots R_{l-1}$ are relations between them. The transition probability for a node at step $i$ is given by:

$$p(v^{i+1}|v^i, P) = \begin{cases} \frac{1}{|N_{t+1}(v_t^i)|} & (v^{i+1}, v_t^i) \in E, \phi(v^{i+1}) = t+1 \\ 0 & (v^{i+1}, v_t^i) \in E, \phi(v^{i+1}) \neq t+1 \\ 0 & (v^{i+1}, v_t^i) \notin E \end{cases} \quad (2)$$

Each of the above algorithm can also be used for learning edge representations. They are obtained by combining node representations of adjacent nodes using binary operators, for example average or Hadamard product. This allows to use these algorithm for edge-related tasks, such as link prediction (predicting whether two nodes should be connected or not) or edge classification.

## 3. PROPOSED FRAMEWORK

### 3.1 Domain model

Proposed model of diagnostic data is inspired by observations in [4]. Let $G = (V, E, T)$ be a heterogeneous undirected network, with nodes $V$, edges $E$ and node/edge types $T = T_v \cup T_e$. Let $\phi(v)$ and $\phi(e)$ be appropriate type mapping functions for nodes and edges. We define following four types of nodes $T_v = \{D, S, N, W\}$. Each node type represent a category of entities from diagnostic domain. They are:

- $D$ - disease
- $S$ - symptom occurrence
- $N$ - symptom name
- $W$ - symptom value

We also define three types of edges $T_e = \{SD, SN, SW\}$. $SD$ represents a relationship between disease and symptom occurrences including both non-specific, specific and pathognomonic ones. It is interpreted as "symptom $s$ can occur in disease $d$". $SN$ is a relationship between symptom and symptom name. Its semantics means "symptom $s$ has name $n$". Each symptom occurrence is associated with exactly one name. Relationship $SW$ occurs between symptom and symptom value and can be interpreted as "symptom $s$ has value $w$". Each symptom occurrence is associated with at least one value. An example of such network is shown in Figure 1.

Proposed model has several important practical features. First, it represents heterogeneous network, allowing us not only to use homogeneous network-based representation learning frameworks, but also utilize the knowledge of edge type. Second feature is an ease to gather the actual data. It can be gathered in a form of triplets $\langle d, n, w \rangle \in D \times N \times W$. The symptom node $s$ is then created as an intermediate node between these three. Another practical assumption was the compliance of diseases and symptom names with International Statistical Classification of Diseases and Related Health Problems (ICD-10) to minimize the risk of data inconsistency.

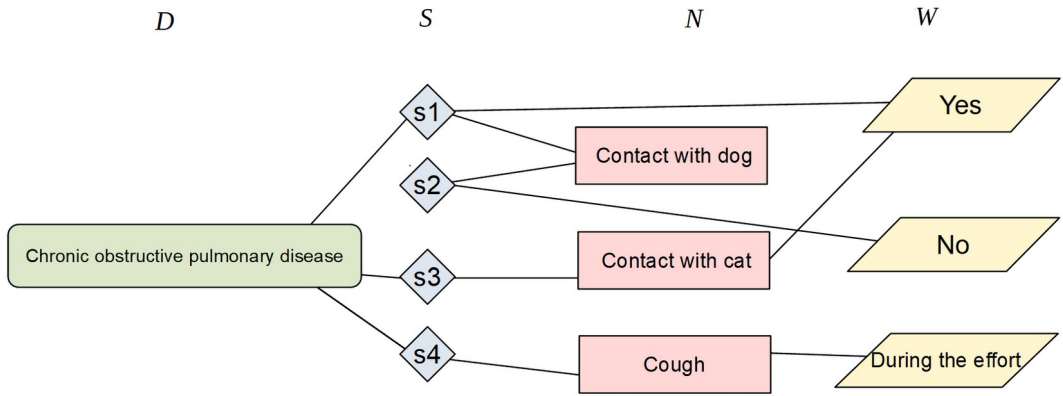

*Figure 1: Example of diagnostic network.*

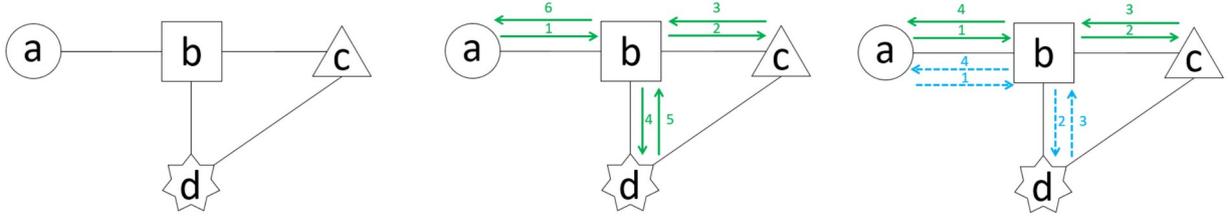

*Figure 2: Illustration of multiple relationships problem. Left: Sample meta-graph consisting of 4 types of nodes. Middle: Meta-path for classical metapath2vec algorithm. Right: meta-paths for proposed algorithm.*

### 3.2  Representation learning algorithm

The original metapath2vec algorithm uses only a single meta-path to generate walks. This may be an issue if relationships that we want to be included does not form a path. An example can be seen in Figure 2. The sample graph contains four types of nodes. Suppose we would like to generate embedding vectors that will incorporate two kind of relationships: between nodes $a$ and $c$ and between nodes $a$ and $d$. Due to the requirement of recursion, the simplest meta-path to include both of must have 7 steps, for example $a-b-c-b-d-b-a$. This may by result in generation of redundant skipgrams as well as may lead to learning undesired features, for example, relationships between nodes $c$ and $d$, and require more iterations to train. We propose a simple modification of the metapath2vec algorithm, by allowing to use multiple shorter metapaths instead of a single one. This way, the desired relationships can be learned by using two metapaths: $a-b-c-b-a$ and $a-b-d-b-a$. Full modified algorithm, which we call multi-metapath2vec, is listed below.

*Algorithm 2: Multi-metapath2vec algorithm*

**Input**: Network $G=(V,E)$, Dimensions $d$, Walks per node $r$, Walk length $l$, Context size $k$, meta-paths $M$
**Output**: node embedding matrix $A \in \mathbb{R}^{|V| \times d}$

Initialize walks to empty
$\rho = \lfloor r/|M| \rfloor$
**for** $i=1 \rightarrow \rho$ **do**
  **for** $v \in V$ **do**
    **for** $m \in M$ **do**
      walk = MetaPathRandomWalk($G,v,l,m$)
      Append walk to walks
    **end**
  **end**
**end**
$A$ = StochasticGradientDescent($k,d,walks$)
**return** $A$

**MetaPathRandomWalk**($G,v,l,m$)
$MP[1]=v$
**for** $i=2 \rightarrow l$ **do**
  $u$ = draw node according to equation (2)
  $MP[i]=u$
**end**
**return** $MP$

## 4. EXPERIMENTS

### 4.1  Experimental setup

Purpose of the experiments is twofold. First, we want to study effect of using of representation learning for diagnostic data by comparing neural networks trained in supervised way with networks of the same architecture, but trained in semi-supervised way, with first layers containing embeddings pretrained with representation learning. Second, we

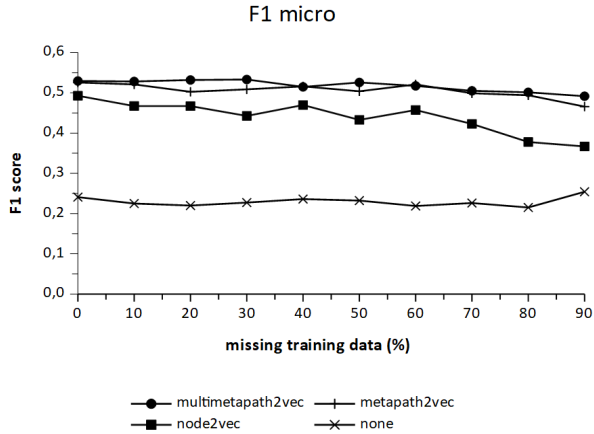
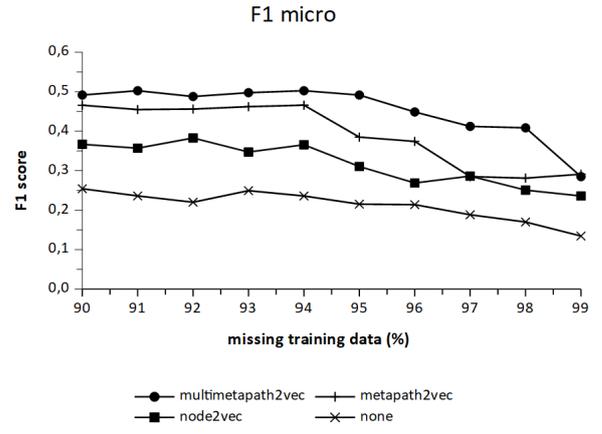
a) b)
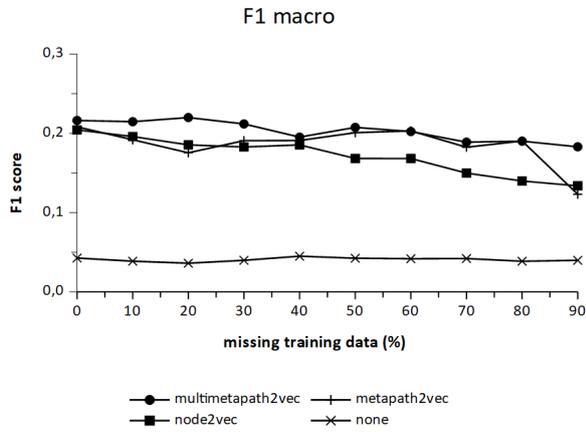
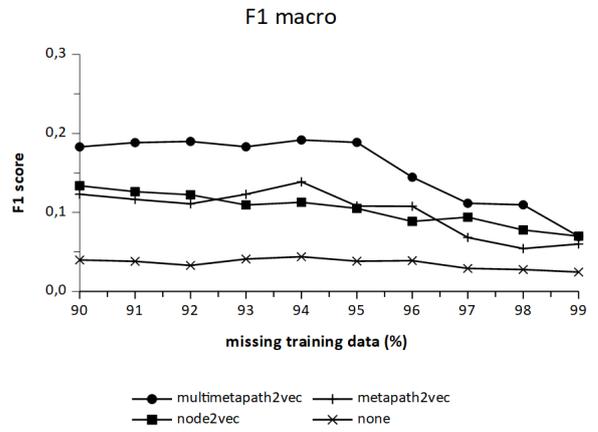
c) d)

Figure 3: F1 scores for node classification task. a) F1 micro score for 0%-90% range. b) F1 micro score for 90%-99% range. c) F1 macro score for 0%-90% range. d) F1 macro score for 90%-99% range.

want to compare the proposed algorithm with other representation learning algorithms. We study two practical tasks involving medical diagnostic data: grouping of symptoms and diseases according to ICD-10 taxonomy and disease prediction.

The dataset used for experiments consists of 14086 triplets $\langle d, n, w \rangle$ which results in an undirected graph consisting of 91 $d$ nodes, 91 $w$ nodes, 728 $n$ nodes, and 1327 $s$ nodes. The data was harvested by a team of physicians as a part of the POIG.02.03.03-00-013/08 project [5] and covers dermatology and pulmonology areas.

Four methods were used to create embedding layers. Each embedding layer was represented by the matrix of the size $d \times k = 2238 \times 100$ with elements sampled uniformly from $[-0.05, 0.05]$ range. Following methods were used:

- No pretraining. The embedding layer was only initialized with random values.

- node2vec: an original implementation, with parameter values $p=1$, $q=1$, $d=2238$, $k=100$, $r=10$, $l=80$.

- metapath2vec: a custom implementation based on the original code, with parameters values the same as above ($p$ and $q$ were not used) and with a single meta-path $d-s-n-s-w-s-d$.

- multi-metapath2vec: a custom implementation, with parameters values the same as

above and with two meta-paths: $d-s-n-s-d$ and $d-s-w-s-d$.

Embedding layers were pretrained with 1 million skip-gram pairs generated using above methods. A single epoch of RMSProp algorithm was used for the training, with default hyper-parameter values and binary cross-entropy as a loss function.

*4.2    Case study: ICD-10 classification*

The task is to classify disease $(d)$ or symptom name $(n)$ nodes according to the subgroup in ICD-10 classification. For example, disease 'Other atopic dermatitis' (ICD code L20.8) is assigned to the subgroup 'L20-L30 Dermatitis and eczema'. The input vector consists of a single node whereas the output vector is onehot-encoded subgroup label. The full dataset contains a total 43 classes – 33 classes of symptom names and 10 classes of diseases. To make the task non-trivial, a certain percentage of the training data is not used. The network should therefore rely on the knowledge about neighbor nodes, in a form of embedding layer, in order to make correct classification. We analyze two ranges of such incompleteness: from 0% to 90% (with step 10%) and from 90% to 99% (with step 1%) of missing data.

The neural networks used in experiments are two-layer feed-forward networks. The first layer is a pretrained embedding layer. The second layer contains 43 neurons with sigmoid activations. Networks are trained with 10 epochs of RMSProp algorithm and binary cross-entropy loss. Cross-validation is used for model evaluation, with $k=10$ and two metrics: F1 micro and macro.

The results for respective ranges and metrics are presented in Figure 3. Pretrained embeddings give significant performance boost compared to the network without pretrained embeddings. Of the representation learning methods, node2vec obtained the worst values, being outperformed by both metapath2vec and multi-metapath2vec for most of the ranges. However, for the 97-99% of missing data, one can observe a rapid performance decrease of metapath2vec. On the other hand, multi-metapath2vec is steadily the best method in terms of both metrics. The performance gap between multi-metapath2vec and other methods, while relatively small for the 90-99% range, becomes particularly visible for the missing data range 90-98%. On the average, multi-metapath2vec obtains 8 % higher F1 micro score than metapath2vec (25% for F1 macro score) and 220 % higher F1 micro score compared to non-pretrained network (471% for F1 macro score).

*4.3    Case study: disease prediction*

In this case we aim to predict whether a set of symptoms is caused by a specific disease. While it represents real-life task of diagnosis, we generate artificial samples using the proposed network model. Each sample is a modeled as a pair of vector of symptoms and a single disease. The symptoms are selected from associated symptom nodes. Additionally, we use a parameter that incorporates data incompleteness by removing randomly $\alpha\%$ of nodes (along with associated edges) from the graph. The full algorithm is given below.

*Algorithm 3: Algorithm for cases generation for disease prediction*

---

**Input**: Heterogeneous network $G=(V,E,T)$, node mapping function $\phi(v):V \to T_v$, cases per disease $n$, maximum symptoms per case $h$, missing data percentage $\alpha$
**Output**: patient cases $C$

Initialize $C$ to empty

$V'=V$ with $\alpha\%$ nodes removed
$V_d=\{v \in V': \phi(v)=D\}$
$V_s=\{v \in V': \phi(v)=S\}$

**for** $v \in V_d$ do
  **for** $i=1 \to n$ do
    $c$ = GenerateCase($V_s, E, v, h, \alpha$)

---

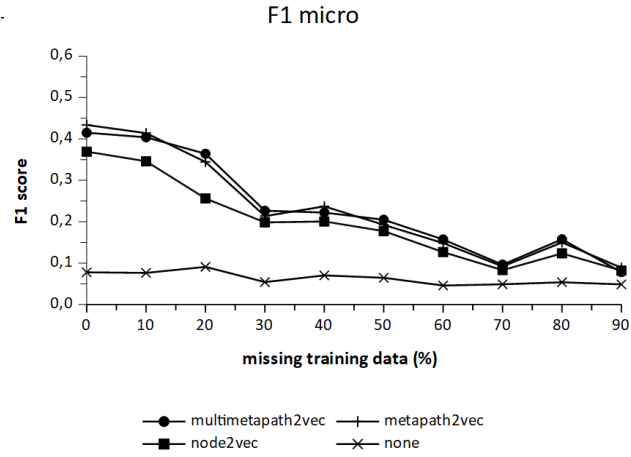
a)

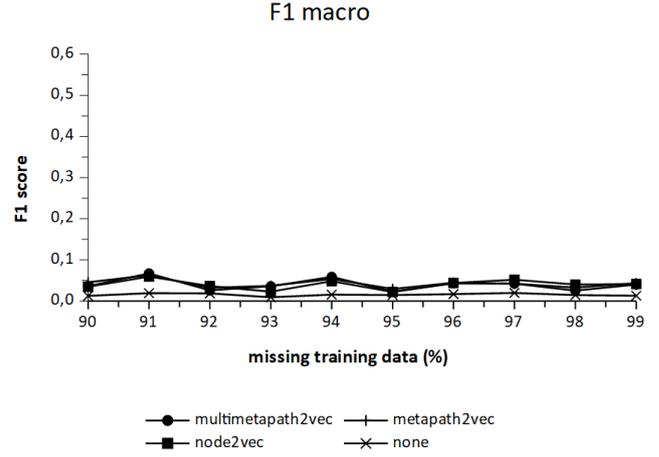
b)

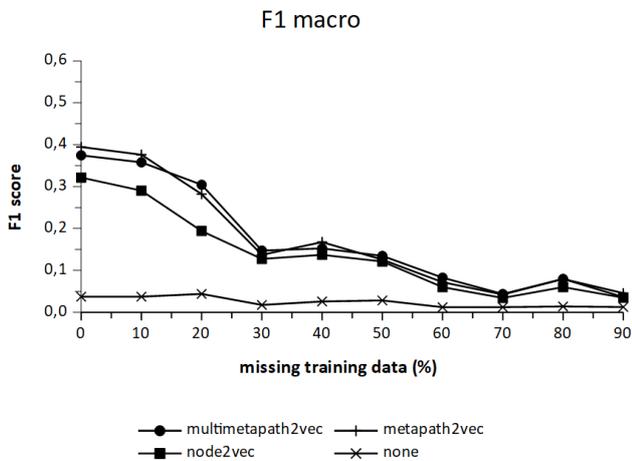
c)

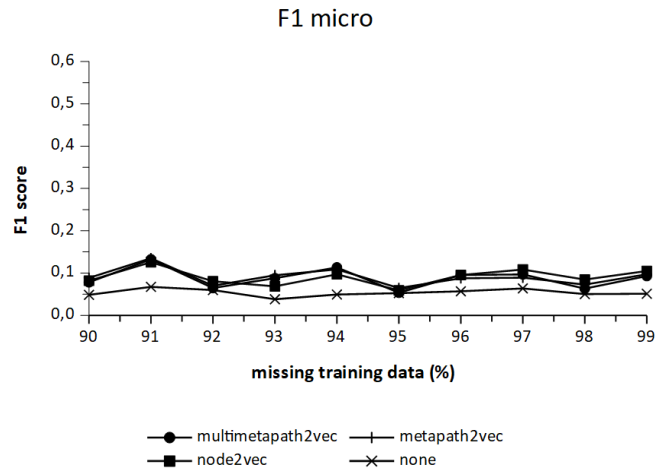
d)

*Figure 4: F1 scores for diagnosis task. a) F1 micro score for 0%-90% range. b) F1 micro score for 90%-99% range. c) F1 macro score for 0%-90% range. d) F1 macro score for 90%-99% range.*

  Append $c$ to $C$
 **end**
**end**
**return** $C$, $C'$

**GenerateCase**( $V_s, E, v, h, \alpha$ )
 $S = \{s \in V_s : \langle v, s \rangle \in E\}$
 $c =$ select at most $h$ elements from $S$
**return** $c$

We generate 10 cases per each disease, using two ranges of $\alpha$ parameter: $0-90\%$ (with step $10\%$) and $90-99\%$ (with step $1\%$). Each case contains at most $h=10$ symptoms. Neural network used for tests are two-layer feed-forward networks with one embedding layer and an output layer with sigmoid activation for one-hot-encoded diseases.

Each network is trained with 10 epochs of RMSProp algorithm, with mean squared error loss. A separate dataset is generated for validation, using the whole graph, with 10 cases per disease. The training and validation is repeated 10 times with different datasets and the performance metrics are averaged.

Test results are presented in Figure 4. Pretrained networks achieved better performance compared to non-pretrained one, which is particularly visible in cases where relatively few data (0-90%) is missing. For values of $\alpha > 90\%$, a benefit of representation learning becomes less significant. Similarly, advantage of metapath2vec and multi-metapath2vec over node2vec is visible when the training graph is mostly untrimmed (0-20% missing nodes). In this case, we have not observed a significant advantage of multi-metapath2vec over non-modified metapath2vec approach.

**5. CONCLUSIONS**

As indicated by test results, using representation learning improves performance of neural networks for medical diagnosis-related tasks such as disease/symptom classification and disease prediction. It follows that the knowledge incorporated in the heterogeneous network model can be efficiently learned and utilized in order to improve machine learning methods used in diagnostic domain to date.

We have shown that heterogeneous network-based metatpath2vec algorithm improves the final performance of the network compared to node2vec. Node2vec requires $p$ and $q$ parameters to be specified that controls influence of structural equivalence and homophily. In most practical cases, relationships between nodes are not only one of them but rather some mixture of both. However, it is difficult to determine the exact proportions before the training. On the other hand, in meta-path-based approach, while we do not need to specify them, the complex relationships can still be incorporated in a form of meta-paths, which are more natural to specify and interpret. Moreover, by allowing to use multiple meta-paths we can avoid traversing unimportant edges that are introduced by the recursion constraint. For certain applications this results in a better performance compared to unmodified metapath2vec, which is especially visible in case of training data shortage.